\documentclass[sigconf]{acmart}

\AtBeginDocument{%
  \providecommand\BibTeX{{%
    \normalfont B\kern-0.5em{\scshape i\kern-0.25em b}\kern-0.8em\TeX}}}

\setcopyright{acmcopyright}
\copyrightyear{2020}
\acmYear{2020}
\acmDOI{10.1145/3394171.3413813}

\acmConference[MM '20]{28th ACM International Conference on Multimedia}{October 12--16, 2020}{Seattle, WA, USA.}
\acmBooktitle{28th ACM International Conference on Multimedia (MM '20), October 12--16, 2020, Seattle, WA, USA.}
\acmPrice{15.00}
\acmISBN{978-1-4503-7988-5/20/10}

\usepackage{soul}
\usepackage{graphicx}
\usepackage{amsmath}
\usepackage{booktabs}
\usepackage{multirow}
\usepackage{latexsym} 
\usepackage[ruled,linesnumbered]{algorithm2e}
\usepackage{makecell}
\DeclareMathAlphabet{\mathcal}{OMS}{cmsy}{m}{n}
\begin{document}

\title{Rethinking Generative Zero-Shot Learning: An Ensemble Learning Perspective for Recognising Visual Patches}





\author{Zhi Chen$^{1}$, Sen Wang$^{1}$, Jingjing Li$^{2}$, Zi Huang$^{1}$}
\affiliation{
  \institution{$^1$The University of Queensland, $^{2}$University of Electronic Science and Technology of China}
}
\email{uqzhichen@gmail.com, sen.wang@uq.edu.au, lijin117@yeah.net, huang@itee.uq.edu.au}

\renewcommand{\shortauthors}{Chen, Zhi, et al.}

\begin{abstract}

Zero-shot learning (ZSL) is commonly used to address the very pervasive problem of predicting unseen classes in fine-grained image classification and other tasks.
One family of solutions is to learn synthesised unseen visual samples produced by generative models from auxiliary semantic information, such as natural language descriptions.
However, for most of these models, performance suffers from noise in the form of irrelevant image backgrounds. Further, most methods do not allocate a calculated weight to each semantic patch. Yet, in the real world, the discriminative power of features can be quantified and directly leveraged to improve accuracy and reduce computational complexity. 
To address these issues, we propose a novel framework called multi-patch generative adversarial nets (MPGAN) that synthesises local patch features and labels unseen classes with a novel weighted voting strategy.
The process begins by generating discriminative visual features from noisy text descriptions for a set of predefined local patches using multiple specialist generative models.
The features synthesised from each patch for unseen classes are then used to construct an ensemble of diverse supervised classifiers, each corresponding to one local patch.
A voting strategy averages the probability distributions output from the classifiers and, given that some patches are more discriminative than others, a discrimination-based attention mechanism helps to weight each patch accordingly.
Extensive experiments show that MPGAN has significantly greater accuracy than state-of-the-art methods.
\end{abstract}



\maketitle
\section{Introduction}

Deep learning has been a blessing and a curse for classification tasks. The performance gains this paradigm has brought are enormous, but they come at the cost of vast and often unrealistic amounts of labelled data. Conventionally, training a traditional classification model requires at least some data samples for all target classes, and deep learning models significantly amplify this issue. Collecting training instances of every class is not always easy, especially in fine-grained image classification \cite{li2018read, li2019discriminative, xie2019srsc}, and therefore much attention has been given to zero-shot learning (ZSL) algorithms as a solution \cite{akata2015label, changpinyo2016synthesized, li2019alleviating, yang2016zero, liu2019attribute, qin2017zero, gao2020zero, shen2020invertible, xie2019attentive}. ZSL expands the classifiers beyond the seen classes with abundant data to unseen classes without enough image samples.

\begin{figure*}[t]
\centering
\includegraphics[width=155mm]{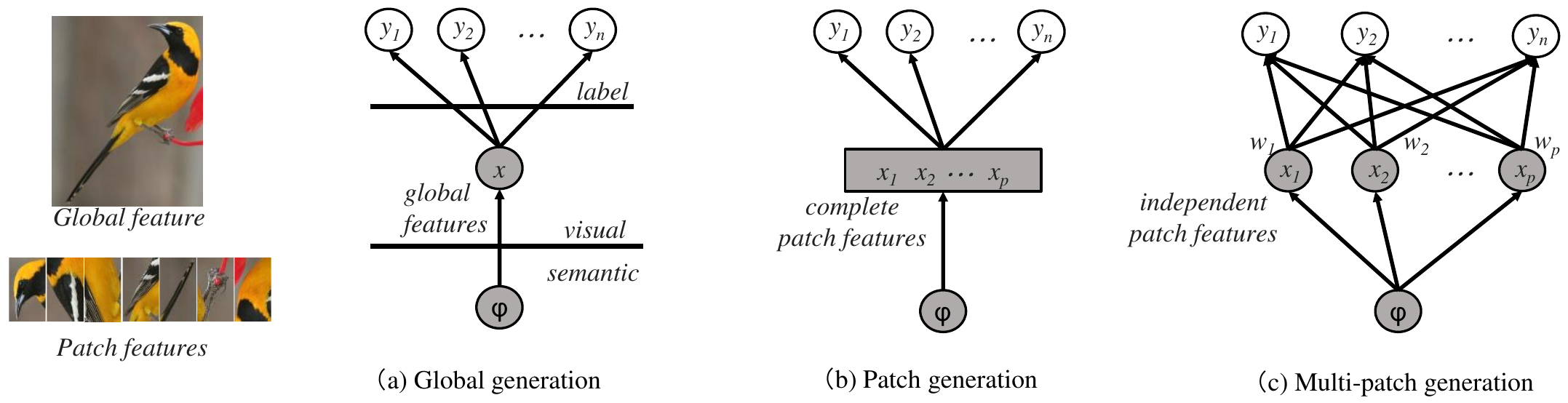}
\vspace{-5pt}
\caption{
A comparison of the global generation, patch generation, and multi-patch generation paradigms. MPGAN is based on multi-patch generation, synthesising features from independent patches and weighting each according to the discriminative power of the patch during classification. Given an unseen class semantic representation $\varphi$, $\textit{x}$ are the synthesized global and $\textit{x}_{p}$ are the $p$-th patch features, $\textit{w}_{p}$ represents an attention weight for $p$-th patch and $y$ denotes the class label.
}
\vspace{-10pt}
\end{figure*}

Generally, ZSL can be categorised into two groups based on how the semantic features are leveraged. These are \textbf{\textit{embedding-based}} methods and \textbf{\textit{generative}} methods. Within the embedding-based methods, there are three mainstream methods to learn embeddings: (1) visual-to-semantic mappings; (2) semantic-to-visual mappings; or (3) latent embedding spaces. Class labels predicted by scoring the compatibility of the visual and semantic features \cite{akata2013label, akata2015evaluation, changpinyo2016synthesized,Region_Graph_Embedding}. By contrast, generative methods deal with ZSL by augmenting data with semantic information and using generative models to "imagine" visual features of the unseen classes. The ZSL task can then be converted into a supervised classification problem \cite{xian2018feature, zhu2018generative}.

Generative ZSL is currently considered to be the state-of-the-art. These approaches allow visual features to be generated for unseen classes with intra-class diversity while retaining inter-class discrimination. A range of techniques have been used to generalize the knowledge learned from seen classes to unseen classes. These include generative adversarial nets (GANs) \cite{goodfellow2014generative}, variational autoencoders (VAEs) \cite{kingma2013auto} and some hybrid variants centred around complicated network design \cite{huang2019generative, li2019leveraging, huang2019generative,chen2019canzsl,xian2019f}. All these methods derive the semantic information needed to generate features from external sources, such as Wikipedia. 

The methods based on generative models either synthesise global visual features extracted from full images, as shown in Figure 1(a) or they synthesise features from multiple patches per image. By contrast, patch generation methods first identify various semantic patches in each image, e.g., head, tail, breast, then they extract the relevant feature from each patch, as shown in Figure 1(b). It is important to note, however, that both patch generation and global generation methods generate all the features for all patches via a single generator.

Despite their promising performance, these approaches have two major limitations in scenarios where fine-grained class distinctions are required. The first is the inability to deal with background noise. The second is the lack of a way to weight the importance of some features over others. 

\begin{itemize}
    \item \textbf{\textit{Background Noise}} \cite{ji2018stacked}: With global generation methods, the extracted visual representations inevitably include some of the background information, which is noise and distracts attention from the discriminative aspects of the local patches. This is particularly problematic in fine-grained classification because objects in two similar classes may only be distinguishable by a few subtle semantic difference. 
    \item \textbf{\textit{Weighting for Importance}}: Most patch generation methods do not include a mechanism for weighting the discriminative power of each patch. Rather, they let the neural network make its own obscure decisions about which features are important and which are not. However, directly leveraging our own knowledge that some features are highly discriminative, while others are shared and therefore entirely irrelevant can bring great benefits to ZSL $-$ most notably, improved accuracy and reduced time complexity. For instance, Crackling Geese and Canada Geese are almost identical, except for the length of the neck. Therefore, in a task to distinguish Crackling geese from Canadian geese, learning to recognise features other than the neck, like the breast or tail, would not only be a waste of time, it would also increase the computational overhead. Focusing on the length of the neck, however, would guide the model toward greater accuracy while reducing the cost of training at the same time.
\end{itemize}

To address these issues, we designed a novel divide-and-conquer framework, called multi-patch generative adversarial nets (MPGAN), that separately synthesises local features for each semantic patch and classifies unseen classes with a novel voting strategy. The framework is illustrated in Figure 1(c). To alleviate the negative impact of background noise, each image is decomposed into multiple patches based on semantic descriptions, which largely eliminates the irrelevant background information. Within each patch, a paired generator and discriminator synthesise the visual feature associated with the patch according to the semantic information. Because each patch has its own GAN, as opposed to the single GAN used by other patch generation methods, the features synthesised by our framework are of inherently better quality in comparison.

To capture the most relevant patterns, a classifier is constructed for each patch that can recognise fine-grained classes. The final classification is made through an ensemble vote on the probability outputs of all the classifiers. Weight assignments based on the discriminative power of the semantic patches ensure the most important patches are brought to prominence. For example, in a task to classify images of birds, the patch containing the "head" feature may be more discriminative than the "tail" patch; therefore, the head would be weighted more highly than the tail.  This weighting, i.e., the level of discrimination, is determined by an attention mechanism that computes both the inter-class and intra-class distances of the features to the class centroid for each patch and dividing the two. With the weightings allocated and the ensemble vote complete, the class labels with the highest confidence in the stacked probability output are taken as the final prediction. 

Overall, our contributions can be briefly summarised as follows:
\begin{itemize}
    \item We propose a divide-and-conquer framework, called MPGAN, that divides an image into multiple semantic patches. Local visual features for each patch are then generated by their own dedicated GAN to improve accuracy with fine-grained ZSL classification tasks. To the best of our knowledge, this is the first attempt to isolate patches and generate visual features with dedicated GANs in generative ZSL.

    \item 
    To exploit prominent local patterns, each patch has its own classifier. Further, a novel attention mechanism calculates weights for each patch based on its discriminative power. These weights help the classification model focus on the patches with more distinctive patterns while skipping those patches with common or identical visual characteristics.
    
    \item Comprehensive experiments and in-depth analyses with two real-world benchmark datasets demonstrate state-of-the-art accuracy by the MPGAN framework in both zero-shot image recognition and zero-shot retrieval.
\end{itemize}
The rest of the paper is organised as follows. We briefly review related work in Section 2. MPGAN is presented in Section 3, followed by the experiments in Section 4. Lastly, Section 5 concludes the paper.

\section{Related Work}
\subsection{Generative ZSL}
A number of generative methods have been proposed for generating visual features in ZSL. GAZSL \cite{zhu2018generative} is a carefully designed generative model based on the improved WGAN \cite{gulrajani2017improved} that synthesises realistic visual features from noisy Wikipedia articles. These new visual centroids provide complementary supervision for regulating the visual feature distribution with inter-class discrimination. CIZSL \cite{elhoseiny2019creativity} directly generates unseen classes by making adjustments, i.e., deviations, to seen classes with the help of a parametrised entropy measure. CADA-VAE \cite{schonfeld2019generalized} is a dual VAE model that finds an intermediate hidden feature space between the visual and attribute spaces. CANZSL \cite{chen2019canzsl} leverages a cycle architecture by translating synthesised visual features into semantic information. A cycle-consistent loss is then applied between the ground truth and the synthesised semantic information. GDAN \cite{huang2019generative} incorporates a flexible metric in the model's discriminator to measure the similarity of features from different modalities. LisGAN \cite{li2019leveraging} considers the multi-view nature of different images and regularises each generated sample to be close to at least one fundamentally representative sample.

Of these methods, GAZSL \cite{zhu2018generative}, CIZSL \cite{elhoseiny2019creativity}, CANZSL \cite{chen2019canzsl} all extract visual features from multiple patches like our framework. However, the patch features in these methods are concatenated as global visual representations, then handled in the same way as global generation methods, thus, neglecting the distinct discriminative information in different patches. By contrast, MPGAN computes a weight for each patch, which results in substantially greater accuracy.

\vspace{-5pt}
\subsection{Fine-grained Image Generation}
A recent trend, especially with multi-stage approaches, is to progressively synthesise vivid fine-grained images at the pixel level using multiple generative modules \cite{zhang2017stackgan,zhang2018stackgan++,xu2018attngan,chen2019cycle}. The process with StackGAN \cite{zhang2017stackgan}, for example, involves sketching the basic shape and colours of an object given a text description in Stage 1, then using the text descriptions and the results from the previous stage as inputs to generate high-resolution images with photo-realistic details in Stage 2. alignDRAW \cite{mansimov2015generating} generates images from a natural language description and considers fine-grained information by iteratively drawing image patches on a canvas, while attending to the relevant words in the caption. TextureGAN \cite{xian2018texturegan} generates plausible fine-grained textures in deep image synthesis with a local patch sampler.

Unlike these pixel-level image generation methods, ZSL for classification tasks only involves feature-level generation. The aim is to generate an image that can be visually interpreted while generating visual features that can directly provide discrimination power for better classification outcomes. In pursuit of better discrimination among generated visual features, MPGAN therefore synthesises multiple discriminative patch features in parallel.
\begin{figure*}[t]
\centering
\includegraphics[width=155mm]{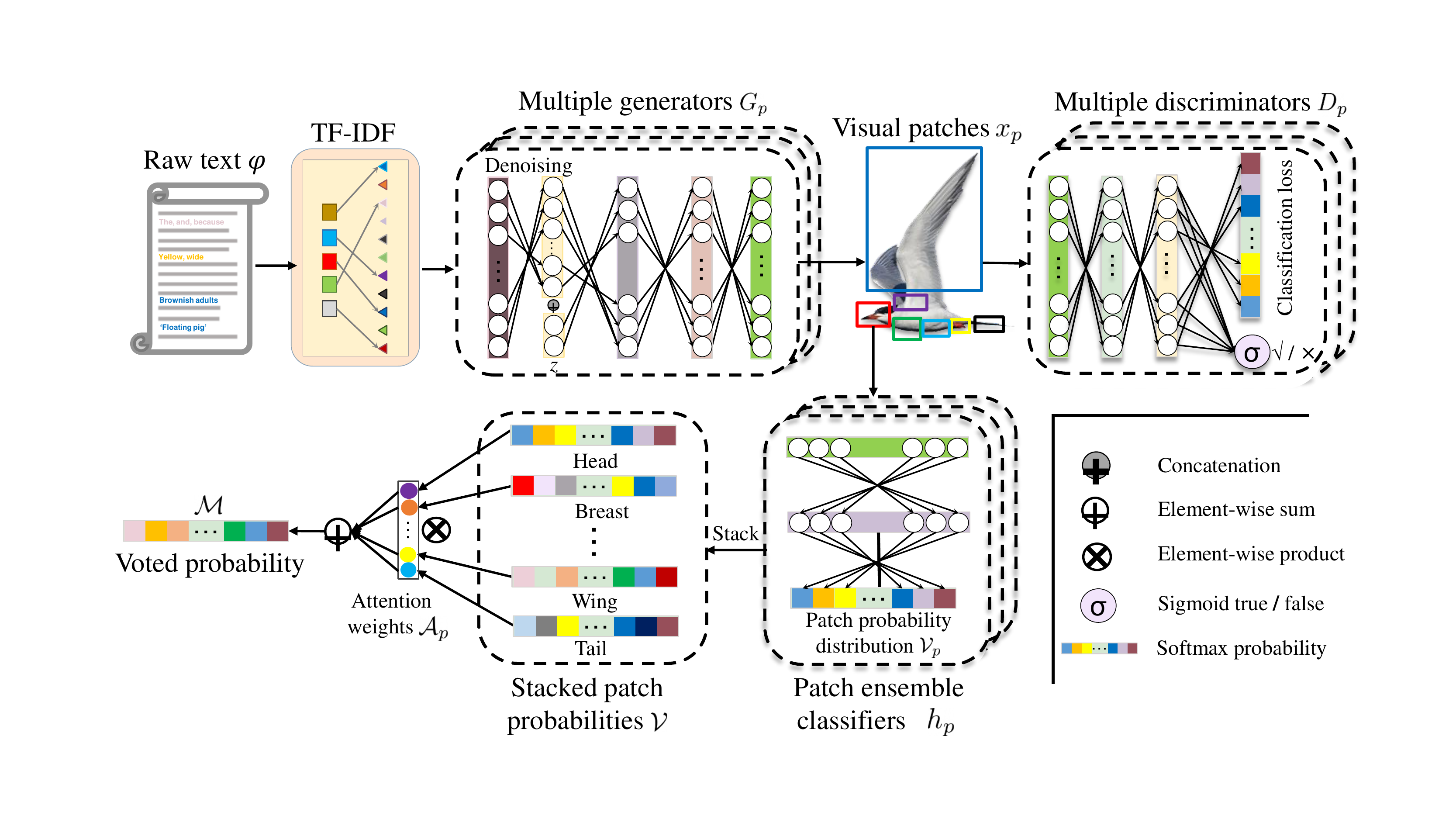}
\vspace{-5pt}
\caption{
Overview of the MPGAN framework. 
MPGAN is a divide-and-conquer framework that divides images into one patch per feature and generates local visual features for each patch with multiple GANs, each dedicated to one patch. 
After the denoised semantic features are concatenated with some random noise, they are used to generate visual features for a patch. The generated features are either fed into the discriminator during the GAN training or used to train the voting classifiers. The output of each patch's classifier is a softmax probability multiplied by an attention weight based on the features' discrimination. The winning probability in the vote is taken as the classification result.
}
\vspace{-10pt}
\end{figure*}
\vspace{-5pt}

\subsection{Attention Mechanisms in ZSL}
In recent years, attention mechanisms have boosted the performance of many deep learning tasks. In ZSL, attention helps to break through the limitations imposed by noisy global features. More specifically, it allows the classification model to attend to the most discriminative areas, effectively eliminating background noise.

In this vein, S2GA \cite{ji2018stacked} is an end-to-end framework with a stacked semantically-guided attention mechanism that progressively guides the visual features toward generated attention weights for different local features. Notably, S2GA integrates both the global visual features and the weighted local discriminant features to represent an image by allocating different weights for different local regions based on their relevance to the semantic features of the class. SGMA \cite{zhu2019semantic} is a localised semantically-guided multi-attention mechanism directed toward global features. A multi-attention loss encourages compact and diverse attention distribution by applying geometric constraints over attention maps. The AREN model \cite{xie2019attentive} discovers multiple semantic parts of images guided by an attention mechanism and the compatibility loss. The model is also coupled with a paralleled network branch to guarantee more stable semantic transfer from the perspective of second-order collaboration.

The conventional attention techniques incorporated into these approaches typically have a high time complexity during training and fail to directly attend to the most critical areas. Hence, MPGAN incorporates a novel discrimination-based attention mechanism that operates as an offline algorithm with pre-computed attention weights.


\vspace{-5pt}
\section{Methodology}
This section begins with the problem formulation and notations. MPGAN is outlined next, followed by the discrimination-based attention mechanism and the ensembled voting strategy.

\vspace{-10pt}
\subsection{Problem Formulation and Notations}
Consider a dataset where the classes are split into two sets - a seen set $\mathcal{S}$ and an unseen set $\mathcal{U}$.Thus, the seen class labels $\mathcal{Y}^{\mathcal{S}}$ in $\mathcal{S}$ and the unseen class labels
  $\mathcal{Y}^\mathcal{{U}}$ in $\mathcal{U}$ are disjoint, \textit{i.e.,} $\mathcal{Y}^{\mathcal{S}}$ $\cap$
 $\mathcal{Y}^{\mathcal{U}}$ = $\varnothing$. The semantic descriptions of the seen and unseen classes are available as
 text embeddings in the semantic space, denoted as $\varphi^{s}$ and $\varphi^{u}$, respectively. Each class label $y$ in
  either the seen or unseen set corresponds to a text embedding $\varphi_{y}^{s}$ or $\varphi_{y}^{u}$. Given $N^{s}$
      labeled training instances $\{(x_{n}^{s},\; y_{n}^{s})\}_{n=1}^{N^{s}}$ in $\mathcal{S}$, where $x_{n}^{s}$ denotes
       the visual features of the n-th image and $y_{n}^{s}$ is the corresponding class label. Each $x_{n}^{s}$ consists of
        $P$ patches, of which each is represented as $x_{n,p}^{s}$, where $p=1,\cdots,P$. The aim of this task is to build a
         model that can accurately classify visual features for unseen classes by learning to generate visual features from
          semantic information. There are $N^{u}$ testing instances $\{(x_{n}^{u},\;y_{n}^{u})\}_{n=1}^{N^{u}}$ in
           $\mathcal{U}$, where $x_{n}^{u}$ are the visual features of the $n$-th unseen testing sample and $y_{n}^{u}$ is
            the unseen class label.

\vspace{-10pt}
\subsection{Generating Visual Patches}
The MPGAN framework is shown in Figure 2. The basic specialist   module for generating each visual patch is a generative adversarial net (GAN) consisting of a discriminator network D and a generator network G. Each net is independent and generates visual features corresponding to its allocated patch $P$ from semantic information. For example, one of the GANs, consisting of $G_p$ and $D_p$, may be allocated to the head patches and so would be trained to only synthesise the head features.  Unlike existing patch generation methods where one GAN generates all the features for all the patches (e.g. head, back, wing, etc.) \cite{zhu2018generative,chen2019canzsl,elhoseiny2019creativity}, using multiple independent GANs dedicated to   generating just the features for one patch results in far more robust visual features.

We selected WGAN-GP \cite{gulrajani2017improved} as the basic generative block because this method introduces a Wasserstein distance and a gradient penalty specifically to prevent the problem of model collapse during training $–$ a common problem with GANs. Further, given that the generated visual features need to be conditioned on the corresponding semantic information, we incorporated Zhu \textit{et al.}'s GAZSL approach \cite{zhu2018generative}  with its baseline settings into the framework to generate a GAN for each semantic patch. GAZSL is an extended version of an adversarial net with an additional classification loss and a regularisation term to increase inter-class discrimination. Each generator $G_{p}$ takes the Term Frequency-Inverse Document Frequency (TF-IDF) representations extracted from the semantic descriptions as its input. However, not all the text in a description is necessarily relevant to a specific patch. For instance, descriptions of the "Tail" and "Back" features are not relevant to the "Head" feature. This information is treated as noise and removed before synthesising the target feature with a denoising module. The denoising module could be either a fully connected layer or Principal Component Analysis (PCA). For convenience, we have provided a comparison between the two in the experiments section. 

Once the denoised text representations have been prepared, they are concatenated with some random noise $z \in \mathbb{R}^Z$ sampled from a Gaussian distribution. These  concatenated representations are then used as the input for two  subsequent fully connected layers,   followed by the activation functions. The objective of the generator $G_{p}$ of the $p$-th patch is defined as
\begin{equation}
\begin{gathered}
  \mathcal{L}_{G_{p}} = - \mathbb{E}_{z \sim p_{z}}[D_{p}(G_{p}(\varphi^{s},z,\theta_{p}),\delta_{p})]   \\ +  \mathcal{L}_{cls_{p}}(G_{p}(\varphi^{s},z,\theta_{p})) + \lambda_{p}||G_{p}(\varphi^{s},z,\theta_{p}) - \hat{x}_{p,y}^{s}||^{2} ,
\end{gathered}
\end{equation}
where $D_{p}$ represents the Wasserstein loss for the $p$-th patch generation, and $\mathcal{L}_{cls_{p}}$ is a classification loss to ensure the inter-class discrimination of the synthesised visual features. $\theta_{p}$ the parameter for the $p$-th patch generator, and $\delta_{p}$ is the parameter for the discriminator. The last term is the visual pivot regularisation that aims to push the generated samples towards the corresponding visual centre of cluster $y$ and patch $p$ in the Euclidean space, denoted as $\hat{x}_{p, y}^{s}$. $\lambda_p$ is the weight for the visual pivot in the $p$-th patch.

The discriminator $D_{p}$ for the $p$-th patch then takes the synthesised visual features as input to calculate both the probability that the features are real as well as the classification loss. Specifically, each discriminator consists of three fully connected layers. The first layer encodes the visual features of the patch, followed by a ReLU activation function. The other two fully connected layers take the encoded visual features and calculate the discriminator loss and the classification loss, respectively. The discriminator loss is a basic loss that distinguishes how realistic the input patch features are. The classification loss is to distinguish the synthesised visual features between classes, which is needed to be able to explicitly classify each feature into a corresponding class. The objective function of the discriminator $D_{s}$ formulated as
\begin{equation}
\begin{gathered}
  \mathcal{L}_{D_{p}} =  \frac{1}{2}(\mathcal{L}_{cls_{p}}(G_{p}(\varphi^{s},z,\theta_{p}))
  +  \mathcal{L}_{cls_{p}}(x^{s}_{p})) +  \beta_{p} \mathcal{L}_{GP_{p}} \\ +  \mathbb{E}_{z \sim p_{z}}[D_{p}(G_{p}(\varphi^{s},z,\theta_{p}),\delta_{p})]  -  \mathbb{E}_{x \sim p_{data}}[D_{p}(x^{s}_{p},\delta_{p})] ,
\end{gathered}
\end{equation}
where the first two $cls_{p}$ represents the classification losses - one for generated patch features and the other for the real patch features. $\mathcal{L}_{GP_{p}}$ is the gradient penalty used to stabilize GAN training, and $\mathcal{L}_{GP_{p}}$ is a Lipschitz constraint, $\mathcal{L}_{GP_{p}} = (||\triangledown_{\tilde{x}^{s}_{p}} D_{p}||_{2} - 1)^2$, where $\tilde{x}_{p}^{s}$ is a linear interpolation between the fake features $\hat{x}^{s}_{p}$ and the real features $x^{s}_{p}$ in the $p$-th patch, and $\beta_p$ is the weight of the gradient penalty in the $p$-th patch. The last term is the Wasserstein loss for the synthesised and real input visual features.

\vspace{-10pt}
\subsection{Ensemble Prediction for Unseen Patches}
With the model trained to generate visual features from semantic information, the next step is to generate realistic visual features for the unseen classes from the semantic information. Once complete, supervised classifiers can be trained to predict labels for each patch.

However, because the overarching recognition task covers all patches, there needs to be a strategy to improve global classification performance by integrating the output probabilities from each patch into the final prediction. A typical approach is a voting strategy over all the individual classification results, but we want to give priority to the more discriminative patches so, as an initial implementation, we incorporated a traditional self-attention mechanism as a weighting method. However, preliminary experiments showed that self-attention did not significantly affect the results of our framework. Considering that the patches are already determined during generation, we then replaced the self-attention mechanism with a straightforward weighted voting technique, reducing the complexity of the framework in the process.
\vspace{-10pt}

\subsubsection{\textbf{Discriminative Attention}}
\begin{algorithm}[tbh]

    \caption{Attention weights $\mathcal{A}_{p}$ computation.}
    \label{alg:algorithm-label}
    
      \SetKwInOut{Input}{Input}
      \Input{the seen patch visual features $x_{p,y}^{s}$, y is the
      corresponding class label, $p$ is the $p$-th patch.}
      \textbf{Step 1:} Compute class centroids $\hat{x}_{p,y}^{s}$ for
      each patch:

        \textbf{for} patch p = 1,...,$P$ \textbf{do} \\
            \hspace{10pt} \textbf{for} class y = 1,...,$\mathcal{Y}$ \\ 
             \hspace{20pt}   $\hat{x}_{p,y}^{s} \leftarrow 
             \frac{1}{N^{y}}\sum_{i}^{N^{y}}x_{p,y}^{s(i)}$,  class y has $N^{y}$ samples \\
        
        \textbf{Step 2:} Compute inter-class and intra-class distance
            $\mathcal{D}_{inter}, \mathcal{D}_{intra}$  for each patch: \\
            
        \textbf{for} patch p = 1,...,$P$ \textbf{do} \\
            \hspace{10pt} \textbf{for} class y = 1,..., $\mathcal{Y}$ \textbf{do} \\
            \hspace{20pt}    $\mathcal{D}_{intra}^{p,y} \leftarrow
            \frac{1}{N^{y}}\sum^{N^{y}}_{i=1} ||x_{p,y}^{s(i)} -
            \hat{x}_{p,y}^{s}||_{2}$ \\
    
                \hspace{20pt} \textbf{find} the nearest centroid to
                $\hat{x}^{s}_{p,y}$ from other classes \textbf{do} \\
                  \hspace{35pt} $\mathcal{D}_{inter}^{p,y} \leftarrow ||\hat{x}_{p,l}^{s} - \hat{x}_{p,y}^{s}||_{2}, (l\neq y)  $ \\
        \textbf{Step 3:} Compute the inter intra-class distance ratio: \\
        \hspace{10pt} $\mathcal{R}_{dis} \leftarrow \frac{\mathcal{D}_{inter}}{\mathcal{D}_{intra}}$ (element-wise division) \\
        \textbf{Step 4:} Compute the discrimination-based attention weights: \\
        \textbf{for} patch p = 1,...,$P$ \textbf{do} \\
            \hspace{10pt} $\mathcal{A}_{p} \leftarrow mean(\mathcal{R}_{dis}^{p})$
        
\end{algorithm}

With the evidence that self-attention had merit, we next need to tackle the issue of weighting the importance of each patch. Inspired by the weighted voting strategy and the failed attention mechanism above, we develop an alternative attention mechanism to weight the importance of each patch based on its discrimination power during the process of predicting labels for unseen classes. These discrimination-based attention weights are computed in a pre-processing step by leveraging the intra- and inter-class difference between the seen classes in the training set for each patch. The computation process is shown in Algorithm 1. By way of summary, first the cluster centroids $\hat{x}^{s}_{p,y}$ for each seen class $y$ in the $p$-th patch are identified. The intra-class distance $\mathcal{D}_{intra}^{p,y}$ for $y$ is then determined by the sum of the Euclidean distance between each visual sample and its associated cluster centroid, and the inter-class distance $\mathcal{D}_{inter}^{p,y}$ is measured according to the distance of the nearest sample from each other class to the class centroid $\hat{x}^{s}_{p,y}$. Next, the inter-class/intra-class distance ratio $\mathcal{R}_{dis}$ is computed from an element-wise division of $\mathcal{D}_{inter}$ and $\mathcal{D}_{intra}$. Lastly, the mean value of the discrimination ratio for all the classes is yielded as the attention weight $\mathcal{A}_{p}$ for each patch.

With the calculation above, the intra-class distance summarises the dispersity of the class distributions, whereas the inter-class distance reflects how distant the class is from other class distributions. As the inter intra-class distance ratio is the quotient of the inter-class distance and the intra-class distance, the less disperse the classes, the higher the ratio. And, given that the ratio of all classes is averaged for each patch, the higher the average ratio, the more discriminative the patch.


\vspace{-5pt}
\subsubsection{\textbf{Ensemble Prediction}}
One softmax classifier $h_{p}$ is trained for each patch $p$. The patch features inputs into the classifiers first pass through a fully connected layer with the number of unseen classes as the output dimension:
\begin{equation}
\begin{gathered}
  h_{p}(\hat{x}^{u}_{p}) =  \mathrm{softmax}(W_{p}\hat{x}^{u}_{p} + b_{p}),
\end{gathered}
\end{equation}
where $\hat{x}^{u}_{p}$ are the synthesised unseen $p$-th patch features, and $W_{p}$ and $b_{p}$ are the parameters of the fully connected layer. Each classifier $h_{p}$ is independently trained to derive the probability distribution  over the unseen classes, where $\mathcal{V}_{p}$ $\in$ $\mathbb{R}^{1\times Y}$, and $Y$ is the total number of unseen classes $\mathcal{Y}^{u}$. Making a prediction for a patch $p$ is done by simply choosing the class index $y^{u}_{p}$ with the highest probability:
\begin{equation}
\begin{gathered}
  y^{u}_{p} =   \underset{y^{u}_{p} \in \mathcal{Y}^{u}}{\mathrm{argmax}}\mathcal{V}_{p}
\end{gathered}
\end{equation}
However, a single patch cannot accurately recognize an object, we then integrate the probability distributions $\mathcal{V}_{p}$ of the various patches to make the ensemble prediction together with the attention weights $\mathcal{A}$ that we computed above.
Since an object cannot be accurately recognised from a single patch, the probability distributions $\mathcal{V}_{p}$ of various patches are integrated and an ensemble prediction is made together with the attention weights $\mathcal{A}$ computed above.

The probability distribution $\mathcal{V}_{p}$ of each patch is multiplied with the corresponding attention weight $\mathcal{A}_{p}$; thus, the scaled probability distribution $\mathcal{V}$ can be calculated with
\begin{equation}
\begin{gathered}
  \mathcal{M}_{p} =   \mathcal{V}_{p} \bigotimes \mathcal{A}_{p},
\end{gathered}
\end{equation}
where $\bigotimes$ is the element-wise product. Each probability value in the vector $\mathcal{V}_{p}$ is multiplied with the scalar $\mathcal{A}_{p}$ to increase or reduce the probability according to the attention weight. Element-wise addition is then performed on the scaled probabilities $\mathcal{M}_{p}$ for each patch $p$:
\begin{equation}
\begin{gathered}
  \mathcal{M} =   \sum_{p=1}^{P} \mathcal{M}_{p}
\end{gathered}
\end{equation}
The final prediction is made by picking the highest probability from the weighted probability distribution $\mathcal{M}$:
\begin{equation}
\begin{gathered}
  y^{u} =   \underset{y^{u}\in \mathcal{Y}^{u}}{\mathrm{argmax}}\mathcal{M}
\end{gathered}
\end{equation}

\section{Experiments}
\subsection{Datasets}
We select two benchmark ZSL datasets for our experiments: Caltech-UCSD Birds 200 (CUB) \cite{wah2011caltech} and North America Birds (NAB) \cite{NAB}. 
The descriptive statistics are provided in Table 1. 
CUB is a small-scale dataset with 11,788 bird images spanning 200 fine-grained species, whereas NAB is a significantly larger dataset with 48,562 images and 1,011 categories. The NAB dataset also has a class hierarchy with 456 parent nodes and 555 leaf nodes. Elhoseiny et al. \cite{elhoseiny2017link} annotated the two datasets with semantic descriptions from Wikipedia articles. However, descriptions of some species were missing so, in the NAB dataset, these species were merged with other classes, leaving 404 annotated classes. There are two suggested splitting strategies for testing classifications tasks with these datasets. The first is the super-category-shared splitting strategy (SCS), where unseen categories with relatively high relevance to seen categories are chosen to share the same super-class. The other strategy is the super-category-exclusive strategy (SCE), where all classes belong to the same super-class but are split into either seen or unseen categories. Intuitively, zero-shot recognition performance should be better with the SCS strategy than with SCE.

\begin {table}[h]
\caption {Statistics of the two ZSL datasets. }
\vspace{-10pt}
\begin{center}
\scalebox{0.82}{
\begin{tabular}{l|cccccc}
\hline
Dataset& TF-IDF Dim & SCS  & SCE  & \#Images & \#Patches &Patch Dim \\  \hline
CUB  & 7,551     & 150/50 & 160/40 & 11,788   & 7   & 512\\
NAB     & 13,217     & 323/81 & 323/81 & 48,562  & 6 & 512    \\ \hline
\end{tabular}}
\end{center}
\vspace{-10pt}
\end{table} 


Since the ultimate goal of ZSL is highly accurate classification, most algorithms operate directly on ready-to-use features extracted from the images, which, as a bonus, results in significantly reduced computation time. The visual features of the CUB and NAB datasets are extracted from VPDEnet following the settings in \cite{elhoseiny2017link}. The CUB dataset returned seven visual patches: "head", "back", "belly", "breast", "leg", "wing" and "tail". NAB lacks the "leg" patch, so there are only six semantic patches. All patches have 512 dimensions. Notably, the superiority of VPDE-net as a global feature extractor has not yet been proven, so we also choose to test the widely-used CNN framework ResNet101 as a comparison. The results of this experiment are given in Section 4.4. 
All the raw text annotations are tokenised into words with punctuation removed, then processed into TF-IDF representations. CUB's representations have 7,551 dimensions, while NAB's have 13,217 (due to differences in the word counts for each dataset). This level of dimensionality is very high so, to reduce the computation cost, we filter out the irrelevant features and embed the representations into a lower-dimensional space in our method.

\subsection{Evaluation Metrics and Comparison Methods}
The average per-class Top-1 accuracy was used as the evaluation criteria, formulated as:
\begin{equation}
\begin{aligned}
  Acc_{\mathcal{Y}} = \frac{1}{\mathcal{Y}} \sum^{\mathcal{Y}}_{y=1} \frac{\# \  of \ correct \ predictions \ in \ y}{\# \ of \ samples \ in \ y},
\end{aligned}
\end{equation}
where $\mathcal{Y}$ is the number of testing classes. A correct prediction is defined as the highest probability of all candidate classes.

We choose six state-of-the-art algorithms for comparison: 
MCZSL \cite{akata2016multi}, ZSLNS \cite{qiao2016less}, ZSLPP \cite{elhoseiny2017link}, GAZSL \cite{zhu2018generative}  , S2GA \cite{ji2018stacked}, CANZSL \cite{chen2019canzsl}, CIZSL \cite{elhoseiny2019creativity}. MCZSL directly uses partial annotations as strong supervision to extract CNN representations of the semantic information in the test phase. GAZSL and ZSLPP simply generate features based on the detected semantic information during both training and testing. S2GA uses semantically-guided attention to identify relevant features, which are then used to progressively generate an attention map for weighting the importance of different local regions. CANZSL is an extension of GAZSL based on cycle architecture that reveals semantic descriptions from visual features, yielding a significant performance improvement. CIZSL introduces a learning signal inspired by the literature on creativity. The unseen space is explored with imagined class-descriptions and careful deviations from the visual features generated from seen classes are encouraged while allowing knowledge transfer from seen to unseen classes.
It is worth mentioning that the basic GAN module in our framework is simpler than the one in CANZSL. As mentioned, CANZSL is based on a cycle architecture designed to reveal semantic information with high accuracy. Therefore, for a fair comparison with GAZSL, the root of CANZSL, we do not add extra components to the GANs, such as a reverse generator like CANZSL.

\begin {table}[t]
\caption {Top-1 accuracy (\%) on the CUB and NAB datasets for each of the two split setting strategies.}
\vspace{-10pt}
\begin{center}
\begin{tabular}[t]{ | l | c | c | c | c |}
\hline
  \multirow{2}{*}{\hfil Methods} & \multicolumn{2}{c|}{CUB}  & \multicolumn{2}{c|}{NAB} \\
  \cline{2-5}
                  & SCS   & SCE   & SCS   & SCE     \\
    \hline
    MCZSL               & 34.7  & -     & -     & -       \\
    ZSLNS               & 29.1  & 7.3   & 24.5  & 6.8     \\
    ZSLPP               & 37.2  & 9.7   & 30.3  & 8.1     \\
    GAZSL               & 43.7  & 10.3  & 35.6  & 8.6     \\
    S2GA                & 42.9  & 10.9  & 39.4  & 9.7       \\
    CANZSL              & 45.8  & 14.3  & 38.1  & 8.9    \\
    CIZSL               & 44.6  & 14.4  & 36.6  & 9.3 \\
  \hline
  MPGAN(ours) & \textbf{48.2} & \textbf{15.5} & \textbf{41.2} & \textbf{10.0} \\
\hline
\end{tabular}
\end{center}
\vspace{-15pt}
\end {table}

\vspace{-10pt}
\subsection{Zero-Shot Learning Results}
Our first experiment test general ZSL classification for each of the methods using the two recommended splitting strategies. We run each experiment multiple times, reporting the average classification accuracy for each dataset and each strategy. The results, shown in Table 2, attest that MPGAN was the most accurate of all the methods with both datasets and both strategies. On CUB, the performance improvement over the next best method, CANZSL, is 2.4\% (SCS) and 1.1\% (SCE) over CIZSL. On NAB, the improvement is 1.8\% (SCS) and 0.3\% (SCE) over S2GA. 
We made several other observations from the results as follows. (1) Accuracy with the SCE splitting strategy is significantly lower than with SCS, which indicates that the relevance of the unseen classes to the seen classes probably has a significant influence over the model's generalisation ability among semantic and visual modalities. (2) Even though S2GA uses both global visual features and those from patches to generate the attention feature map, MPGAN was more accurate using the patch features alone. (3) MPGAN delivers a performance improvement on the CUB dataset of 4.5\% (SCS) and 5.2\% (SCE) over GAZSL and 5.7\% (SCS) and 1.4\% (SCE) on NAB. This is significant because GAZSL can be thought of as a basic version of our MPGAN framework. Overall, the results show a substantial improvement over the current state-of-the-arts, verifying MPGAN's accuracy with ZSL classification tasks.

\subsection{Comparison with Global Generation}
The next set of experiments is designed to evaluate the quality of MPGAN's generated patch and global features. We first fine-tune a ResNet101 model on a training set consisting of the seen classes from CUB. We then use the trained model without the last fully connected layer to extract 2048-dimensional visual features of both the seen and unseen classes. With TF-IDF representations, the global visual features only reach around 10.5\% accuracy with SCS and 4.3\% with SCE. Other ZSL methods that consider global visual features are capable of much better performance – for example, \cite{li2019leveraging,xian2019f,schonfeld2019generalized}. This unsatisfactory result is mainly because the semantic attributes in the dataset are manually annotated by domain experts and are therefore very clean. However, taking that amount of time to label the instances would seldom be practical in zero-shot settings. Hence, we simply extract the semantic information from Wikipedia articles.

\begin {table}[t]
\caption {
The effect of the different components of MPGAN with a zero-shot classification task. The results show the accuracy (\%) with the SCS splitting strategy on the CUB and NAB datasets.
}

\begin{center}
\begin{tabular}[t]{ |l | c | c | c | c |}
\hline
   & \multicolumn{2}{c|}{CUB}  & \multicolumn{2}{c}{NAB} \\
  \hline
  Methods             & Dyn-FC   &  PCA   & Dyn-FC  & PCA      \\
\hline
MP-only             & 44.7  & 43.2   & 37.5  & 35.8       \\
MP-MC-only          & 45.5  & 44.3   & 38.1  & 37.0    \\
MP-MC-MCls-only     & 46.8  & 46.2   & 39.5  & 40.5     \\

  \hline
  MPGAN            & \textbf{48.2}  &\textbf{47.7}  & \textbf{40.2}   &\textbf{41.2}    \\
\hline
\end{tabular}
\end{center}
\vspace{-15pt}
\end {table}

\subsection{Ablation Study}
In this ablation study, we test different variants of MPGAN's overall architecture. These included: MP-only $–$ generating patch features with global visual centroids; MP-MC-only $–$ generating features from multiple patches with multiple centroids; MP-MC-MCIs-only $–$ generating features from multiple patches with multiple centroids and predicting the labels via a single classifier; and MPGAN $–$ the complete method.
The MP-only variant generates visual features for each patch separately, but a global visual centroid is applied to an ensemble of the generated visual patches. MP-only outperforms the GAZSL baseline, which generates visual features for all patches via a single GAN. This result indicates that parallel multi-patch generation is effective. 
The MP-MC-only variant tests how the centroids affect performance. The results of this experiment confirm that the centroids contribute to the discriminative power of the synthesised visual features. 
We also find that the voting strategy with multiple classifiers for each patch prediction, as opposed to MP-MC-MCIs-only with a single classifier, proved to be helpful. Further, there is an obvious improvement when attention weights are incorporated into the voting strategy. 
The specific results for each method appear in Table 3, showing that, as more components are added with each variant, performance gradually improves. Hence, each component makes a contribution to the accuracy of the final prediction.

\vspace{-10pt}
\subsection{Noise Suppression}
Given the TF-IDF representations of text descriptions are high-dimensional (7,551 for CUB and 13,217 for NAB), the time complexity of training/inference can be relatively high if denoising operations involve the entire network. Hence, this experiment is designed to test whether pre-processing the text representations into a lower-dimensional feature space using PCA would deliver the same performance as dynamic denoising through a fully connected layer in the overall model. As there are only 200 classes in the CUB dataset, 200 principal components are the maximum that can be preserved. Obviously, reducing the dimensionality from 7,551 to 200 will result in the loss of some non-trivial information. In contrast, the 404 classes of the NAB dataset mean that significantly more information can be preserved. The performance comparisons between the two noise suppression methods are given in Table 2.


 \begin{figure}[t]
\centering
\includegraphics[width=85mm]{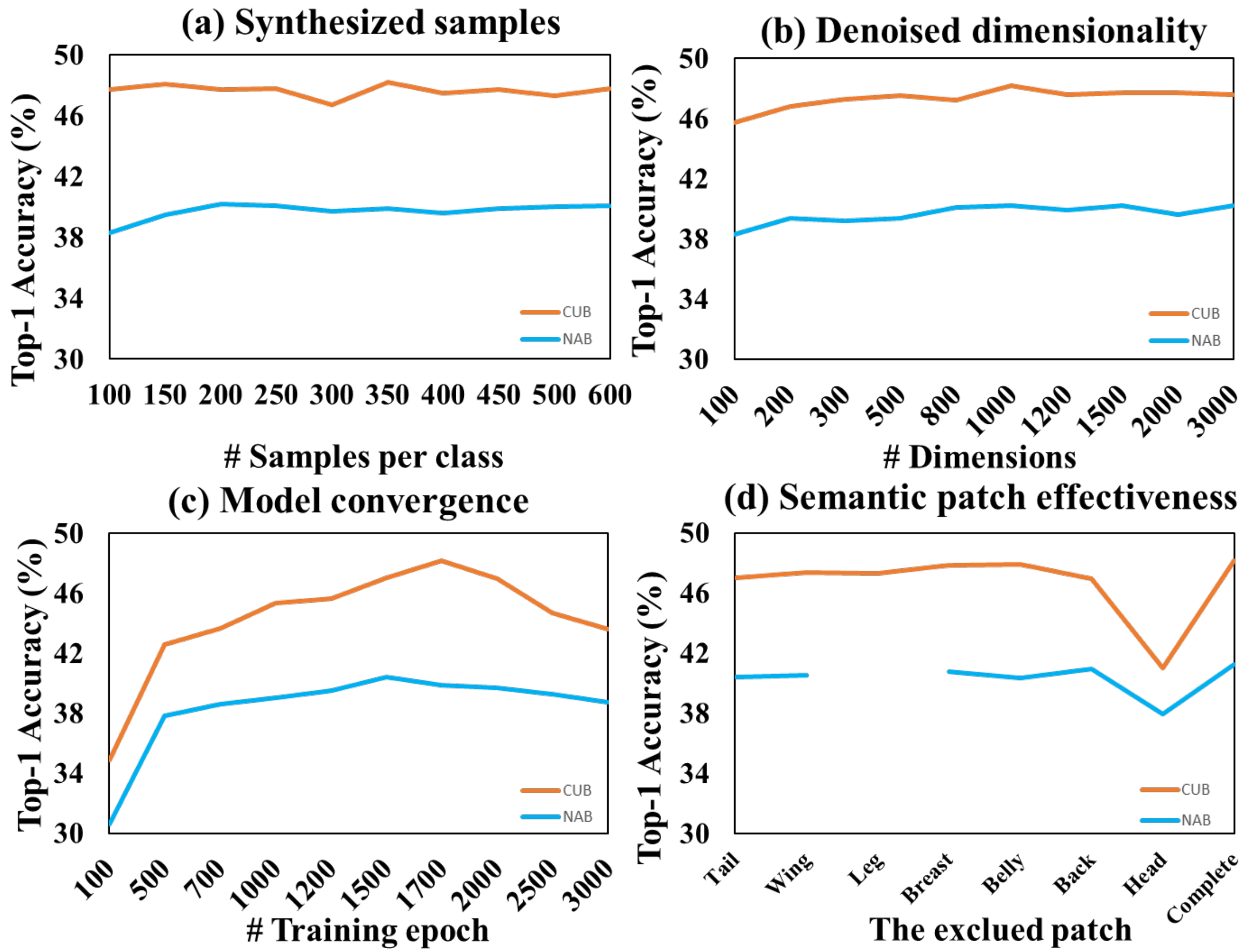}
\vspace{-20pt}
\caption{Parameter sensitivity. Note that the NAB dataset in Fig. 4(d) does not include a "leg" class.}
\vspace{-20pt}
\end{figure}

\begin{figure*}[t]
\centering
\includegraphics[width=175mm]{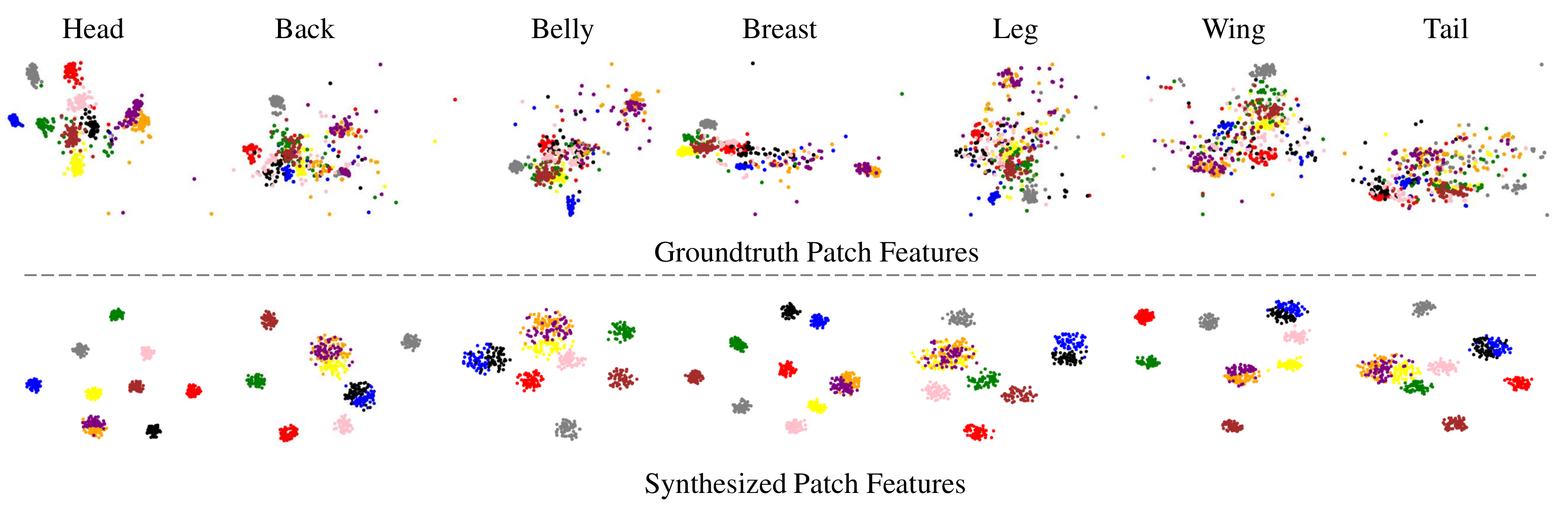}
\vspace{-15pt}
\caption{t-SNE visualisation of patch features from randomly selected unseen classes. The colour represents different class labels. The top row shows the extracted features for training, whereas the second row is the generated patch features.}
\vspace{-10pt}
\end{figure*}

\vspace{-10pt}
\subsection{Parameter Sensitivity}
In a series of experiments, we test the impact of main four hyper-parameters on accuracy. 
Varying the number of generated samples from 100 to 600 in steps of 50, we find little impact on accuracy, as shown in Figure 3(a). Overall, accuracy reaches its peak at around 200 samples for each unseen class on the NAB dataset and 350 on CUB. 
The number of denoised dimensions also has a relatively small impact. Figure 3(b) shows the results of denoising the TF-IDF representations at various levels of dimensionality through a fully connected layer before feeding them into the generative model. The highest accuracy for both datasets occurs at around 1000-dimensions, which would likely apply to any bird recognition task. However, the peak dimensions for other tasks would need to be confirmed in future research. 
The effect of different numbers of training epochs is shown in Figure 3(c). Overall, MPGAN proves to be very stable. After an initial climb, accuracy plateaued at around 500 iterations and only begins to overfit the seen classes at 2000 iterations. 
The last and perhaps most important sensitivity test is the influence of the different patches. The results in Figure 3(d), indicate that MPGAN is the least accurate when the head patch is excluded on both datasets. In a t-SNE analysis, Figure 4 confirms the discrimination of this particular patch. However, despite the importance of the head patch, the best performance occurs when every patch is considered, which demonstrates that each patch is critical. 


\begin {table}[t]
\caption {Comparison of Zero-shot retrieval on the CUB and NAB datasets (mAP (\%)).}
\vspace{-5pt}
\begin{center}
\begin{tabular}[t]{ | l | c | c | c | c | c | c |}
\hline
   & \multicolumn{3}{c|}{CUB}  & \multicolumn{3}{c}{NAB} \\
  \hline
  Methods   & 25\%   &  50\%   & 100\%  &  25\%   &  50\%   & 100\%     \\
\hline
ESZSL       & 27.9      & 27.3      & 22.7      & 28.9      & 27.8      & 20.9      \\
ZSLNS       & 29.2      & 29.5      & 23.9      & 28.8      & 27.3      & 22.1  \\
ZSLPP       & 42.3      & 42.0      & 36.3      & 36.9      & 35.7      & 31.3     \\
GAZSL       & 49.7      & 48.3      & 40.3      & 41.6      & 37.8      & 31.0     \\
S2GA        & -         & 47.1      & 42.6      & -         & \textbf{42.2}      & \textbf{36.6}  \\
CIZSL       & 50.3      & 48.9      & \textbf{46.2}& 41.0   & 40.2      & 34.2 \\
  \hline
MPGAN &\textbf{54.6}  &\textbf{52.2}& 45.1  & \textbf{45.6} & 41.5 & 35.1    \\
\hline
\end{tabular}
\end{center}
\vspace{-15pt}
\end {table}

\vspace{-5pt}
\subsection{t-SNE Analysis}
The t-SNE visualisation in Figure 4 shows the extracted visual features and the synthesised visual features for the different semantic classes with the CUB dataset. The features included in the plot are randomly picked from unseen classes. It is clear that each patch has a different level of discriminative power. For example, the head, as discussed in the previous experiment, is extremely discriminative, as is the leg according to the ground truth. However, relying on the head patch alone will only result in an accuracy of about 35\%. Since the framework is designed to push the generated samples towards the visual centroid of each class during training, the synthesised features are intra-class discriminative. However, due to the lack of discrimination in some patches, the distributions of the synthesised features for some classes significantly overlap. This issue illustrates the reasoning behind the discrimination-based attention mechanism, which mitigates the problem by assigning different weights based on the discriminative power of the patch.

\vspace{-5pt}
\subsection{Zero-shot Retrieval}
To further verify the effectiveness of the MPGAN framework, we also perform the zero-shot retrieval task. In zero-shot retrieval, images of unseen classes must be classified and returned based only on semantic information. We choose mean average precision (mAP) as the evaluation metric, where Precision is calculated as the percentage of correct images retrieved for the class. Table 4 shows the accuracy of each method when retrieving 25\%, 50\% and 100\% of the images for all unseen classes, and Figure 5 illustrates an example of the results for four randomly-chosen classes. The images shown are the five nearest images to the class. The results in Table 4 show MPGAN with the highest accuracy at the 25\% and 50\% levels on the CUB dataset, and at the 25\% level with NAB. The first column of Figure 5 shows a typical image of the bird species. The rest of the columns show the nearest five images in order. Images with green boxes are correctly classified; red boxes are incorrect.

 \begin{figure}[t]
\centering
\includegraphics[width=82mm]{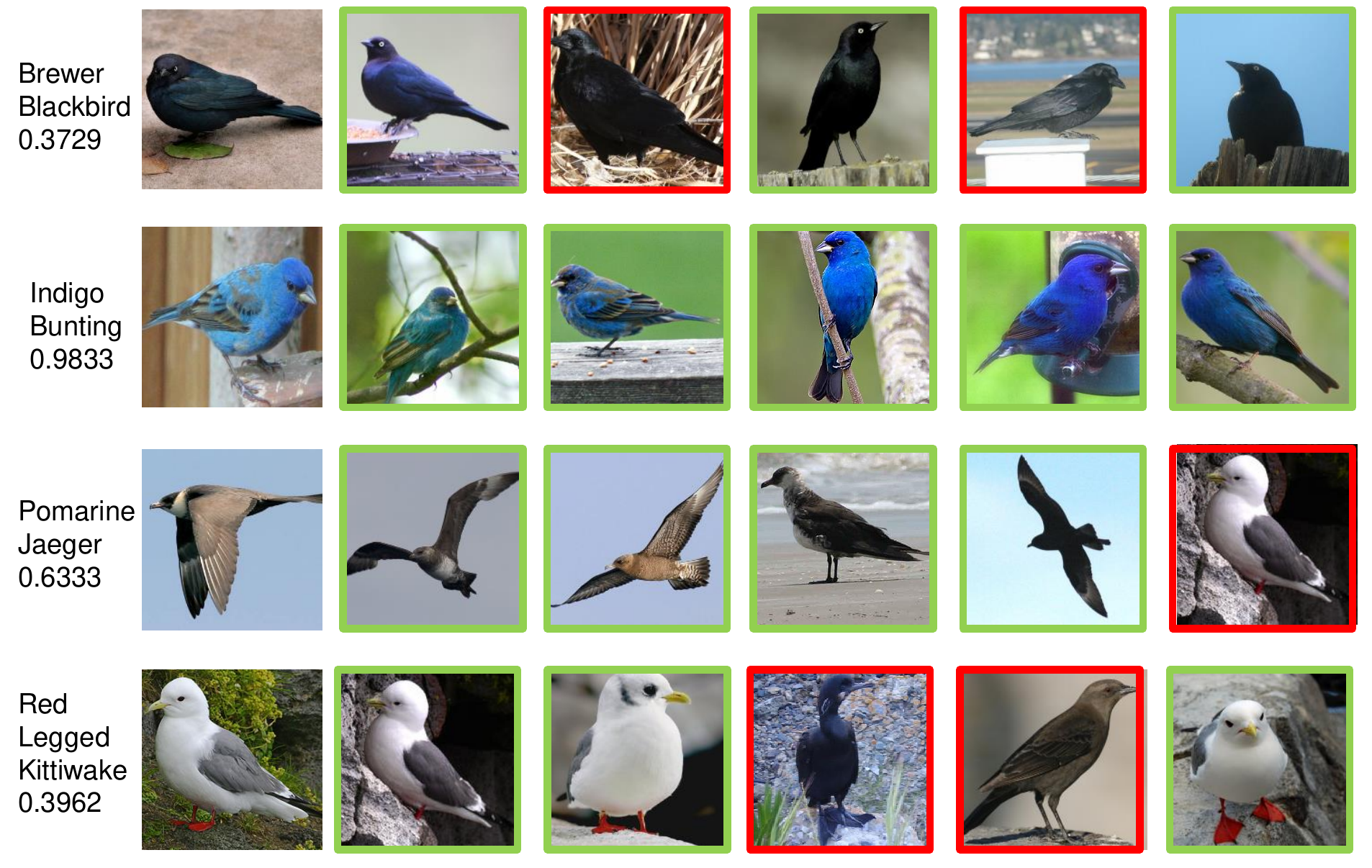}
\vspace{-10pt}
\caption{Visual samples retrieved from the CUB dataset with MPGAN. Correct instances are mark in green; incorrect in red. }
\vspace{-20pt}
\end{figure}

\vspace{-5pt}
\section{Conclusion}
In this paper, we presents a generative paradigm for ZSL that breaks images into a series of fine-grained patches to reveal subtle visual distinctions in features. Called multi-patch generative adversarial networks (MPGAN), the framework decomposes object classification into one classification sub-task per patch. An attention-based voting mechanism then considers the subtle difference between fine-grained patches among classes. An extensive suite of experiments show that MPGAN is a more accurate solution for zero-shot image recognition than seven state-of-the-art approaches with competitive accuracy in zero-shot retrieval. In future work, we intend to generalise MPGAN to other nuanced subject matter, such as flowers or dogs.

\bibliographystyle{ACM-Reference-Format}
\bibliography{sample-base.bbl}

\end{document}